\documentclass{article}
\usepackage{spconf,amsmath,graphicx,hyperref}
\usepackage{amsfonts,amssymb}
\usepackage{xspace}
\usepackage{xcolor}
\usepackage[flushleft]{threeparttable}
\usepackage{pifont}
\usepackage{multirow}
\usepackage{booktabs}
\usepackage{extra_styles}
\newcommand{\vX}{\mathbf{X}}

\usepackage{setspace}   
\usepackage[numbers,sort&compress]{natbib}  

\newenvironment{myequation}{\vspace{-1mm}\begin{equation}}{\end{equation}\vspace{-1mm}}

\title{FastAV: Efficient Token Pruning \\ for Audio-Visual Large Language Model Inference}
\name{Chaeyoung Jung, Youngjoon Jang, Seungwoo Lee, Joon Son Chung}
\address{Korea Advanced Institute of Science and Technology, South Korea 
}              
\begin{document}
%
\maketitle
\begin{abstract}
In this work, we present FastAV, the first token pruning framework tailored for audio-visual large language models (AV-LLMs). While token pruning has been actively explored in standard large language models (LLMs) and vision-language models (LVLMs), its application to AV-LLMs has received little attention, even though multimodal integration substantially increases their token demands. 
To address this gap, we introduce a pruning strategy that utilizes attention weights to identify tokens emphasized at different stages and estimates their importance. Building on this analysis, FastAV applies a two-stage pruning strategy: (1) global pruning in intermediate layers to remove broadly less influential tokens, and (2) fine pruning in later layers considering the impact on next token generation. Notably, our method does not rely on full attention maps, which makes it fully compatible with efficient attention mechanisms such as FlashAttention. Extensive experiments demonstrate that FastAV reduces FLOPs by more than 40\% on two representative AV-LLMs, while preserving or even improving model performance.

\end{abstract}
\begin{keywords}
Audio-visual LLM, LLM token pruning, Multimodal LLM
\end{keywords}
\section{Introduction}
\label{sec:intro}


Recent advances in large language models (LLMs)~\cite{achiam2023gpt, brown2020language, liu2023summary, thoppilan2022lamda} have driven growing interest in extending them to multimodal settings by incorporating vision, audio, and other modalities to tackle more complex tasks.
However, as model sizes and the number of processed tokens grow substantially, both memory consumption and computational cost increase sharply, leading to degraded inference efficiency. To address these challenges, recent studies have explored various strategies to improve model efficiency, ranging from structure-level pruning~\cite{ma2023llm, guo2025slimllm, frantar2023sparsegpt, sun2023simple, zhang2023dynamic} to token-level pruning~\cite{fu2024lazyllm, chen2024image, lin2025boosting, yang2025topv,lee2025token}, all aiming to reduce overhead while preserving model performance.

Several pruning strategies for LLMs have been proposed and can be categorized as post-training, training-free, and inference-time methods. Post-training methods compress pretrained models via structured pruning, including gradient-guided pruning with LoRA recovery~\cite{ma2023llm} and refined importance metrics~\cite{zhang2024plug, guo2025slimllm}. Training-free methods enhance model efficiency without fine-tuning, such as Hessian-based reconstruction~\cite{frantar2023sparsegpt}, activation-preserving criteria~\cite{sun2023simple}, and dynamic sparse masks~\cite{zhang2023dynamic}. 
For inference, LazyLLM~\cite{fu2024lazyllm} dynamically prunes tokens during generation by selectively computing key–value pairs for tokens deemed important, even if they are discarded in earlier layers. 
Extending inference-time token pruning to large vision-language models (LVLMs), recent studies focus on efficient visual token selection. FastV~\cite{chen2024image} prunes visual tokens in later layers based on adaptive attention patterns learned in early layers. VTW~\cite{lin2025boosting} removes mid-to-late visual tokens using attention-sink and information-migration signals. TopV~\cite{yang2025topv} prunes less important visual tokens using a visual-aware cost function.

Unlike the extensive study of token pruning in LLMs and LVLMs, its application to audio-visual LLMs (AV-LLMs) remains underexplored.
AV-LLMs~\cite{chowdhury2024meerkat, sun2024video, tang2025video, cheng2024videollama, ye2024cat, zhan2024anygpt, lu2024unified, han2023imagebind, lyu2023macaw, zhang2023video, chen2023vast, su2023pandagpt, zhao2023chatbridge, han2024onellm, jung2025fork} handle rich multimodal streams of audio, video, and text, leading to an inevitable increase in token counts. However, it is still unclear how many of these tokens are essential for downstream reasoning. To bridge this gap, we present FastAV, a framework that analyzes AV-LLMs to identify key audio-visual tokens and prune less informative ones at inference, enabling efficient computation without compromising performance.

\begin{figure}[t]
  \centering
  \vspace{-2mm}
  \includegraphics[width=0.75\linewidth]{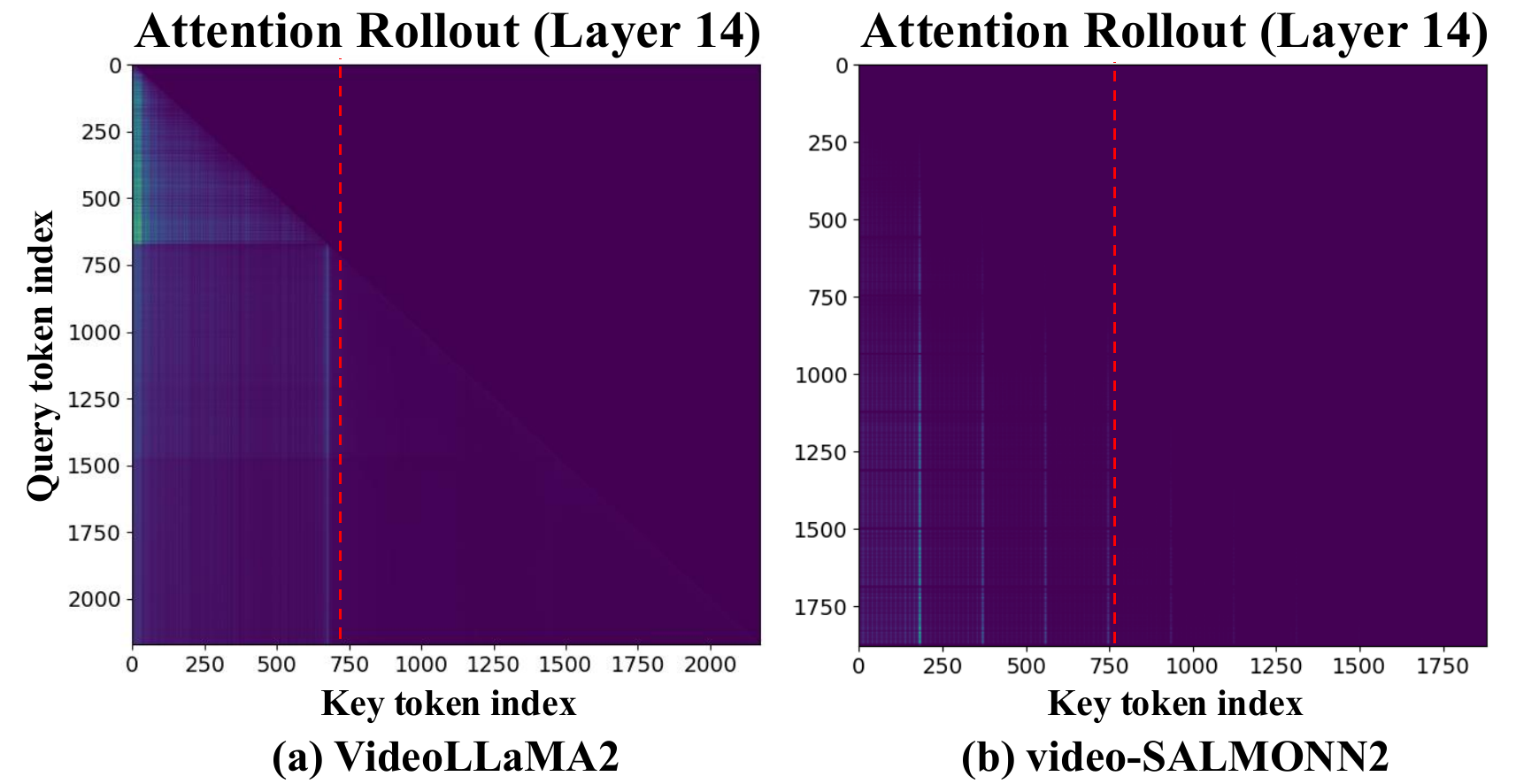}
  \vspace{-4mm}
  \caption{\textbf{Attention rollout at the 14th layer in VideoLLaMA2~\cite{cheng2024videollama} and video-SALMONN2~\cite{tang2025video}.} Accumulated attention concentrates on earlier tokens, highlighting their pivotal role in carrying the most essential information.}
  \vspace{-4mm}
  \label{fig1}
\end{figure}

To gain deeper insights into the behavior of AV-LLMs, we analyze attention rollout~\cite{abnar-zuidema-2020-quantifying}, which accumulates attention across layers to trace token influence. Higher rollout values indicate stronger influence, highlighting critical tokens. We apply this to two representative AV-LLMs, VideoLLaMA2~\cite{cheng2024videollama} and video-SALMONN2~\cite{tang2025video}. 
As shown in Fig.\ref{fig1}, our rollout analysis reveals that after tokens pass through the middle layer (layer 14), attention increasingly concentrates on earlier tokens, particularly on the left side of the red line, forming anchor-like patterns~\cite{huang2024opera} in both models. In the first row of Fig.~\ref{fig2}, we observe that rollout in early layers such as layer 4 remains uniform, whereas by layer 14 it clearly shifts toward earlier tokens. This rollout pattern then persists through deeper layers, including layer 24.

Motivated by this observation, we apply global pruning at the middle layer to roughly remove less influential tokens. Since this stage already eliminates more than half of the tokens, the remaining ones must be pruned with caution to avoid harming model performance. To further increase inference efficiency while preserving important information, we introduce a fine pruning stage. Following prior studies~\cite{song2024hierarchical, jung2025avcd}, we estimate token importance using the attention weights of the last query token, and progressively remove the least important tokens at each subsequent layer based on their contribution to the final answer, further reducing computational costs.

Experimental results show that FastAV reduces FLOPs by over 40\% in two representative AV-LLMs, VideoLLaMA2 and video-SALMONN2, while maintaining or even improving performances across three datasets. 
Furthermore, applying FastAV to both AV-LLMs drastically reduces audio tokens (e.g., from 1,496 to 10 in VideoLLaMA2) without degrading performance, highlighting their essential yet compact role and underscoring the need for efficient strategies to leverage multimodal information, particularly audio.

Our contributions are threefold. First, we present FastAV, the first systematic analysis of the role and impact of audio and video tokens in AV-LLMs. Second, we introduce a two-stage pruning strategy that combines global pruning guided by attention rollout with fine grained pruning based on last query token importance. Third, we show that FastAV reduces FLOPs by more than 40\% while maintaining or even improving model performance, enabling efficient inference.

\section{Method}
\label{sec:method}
\subsection{AV-LLM Inference}
For video data, a visual encoder extracts frame-level features into a sequence of $M$ visual tokens $\vX^{vis} = [\vx_1^{vis}, ..., \vx_M^{vis}]$. Similarly, an audio encoder transforms audio signals into $U$ audio tokens $\vX^{aud} = [\vx_{1}^{aud}, ..., \vx_{U}^{aud}]$. Text inputs, including a question, are tokenized into $E$ textual tokens $\vX^{lang} = [\vx_{1}^{lang}, ..., \vx_{E}^{lang}]$. These modality-specific tokens are concatenated to form a unified sequence $\vX = [\vx_i]_{i=1}^K$, where $K = M + U + E$, which serves as the input to the LLM decoder. 
During inference, the model generates outputs autoregressively, predicting the next token conditioned on all input modalities and previously generated outputs:
\begin{myequation}
\vy_t \sim p(\vy_t|\vX^{vis}, \vX^{aud}, \vX^{lang}, \vy_{<t}),
\end{myequation}
where $\vy_t$ is the token at timestep $t$ and $\vy_{<t}$ are past tokens. 
\begin{figure}[t]
  \centering  \includegraphics[width=1.0\linewidth]{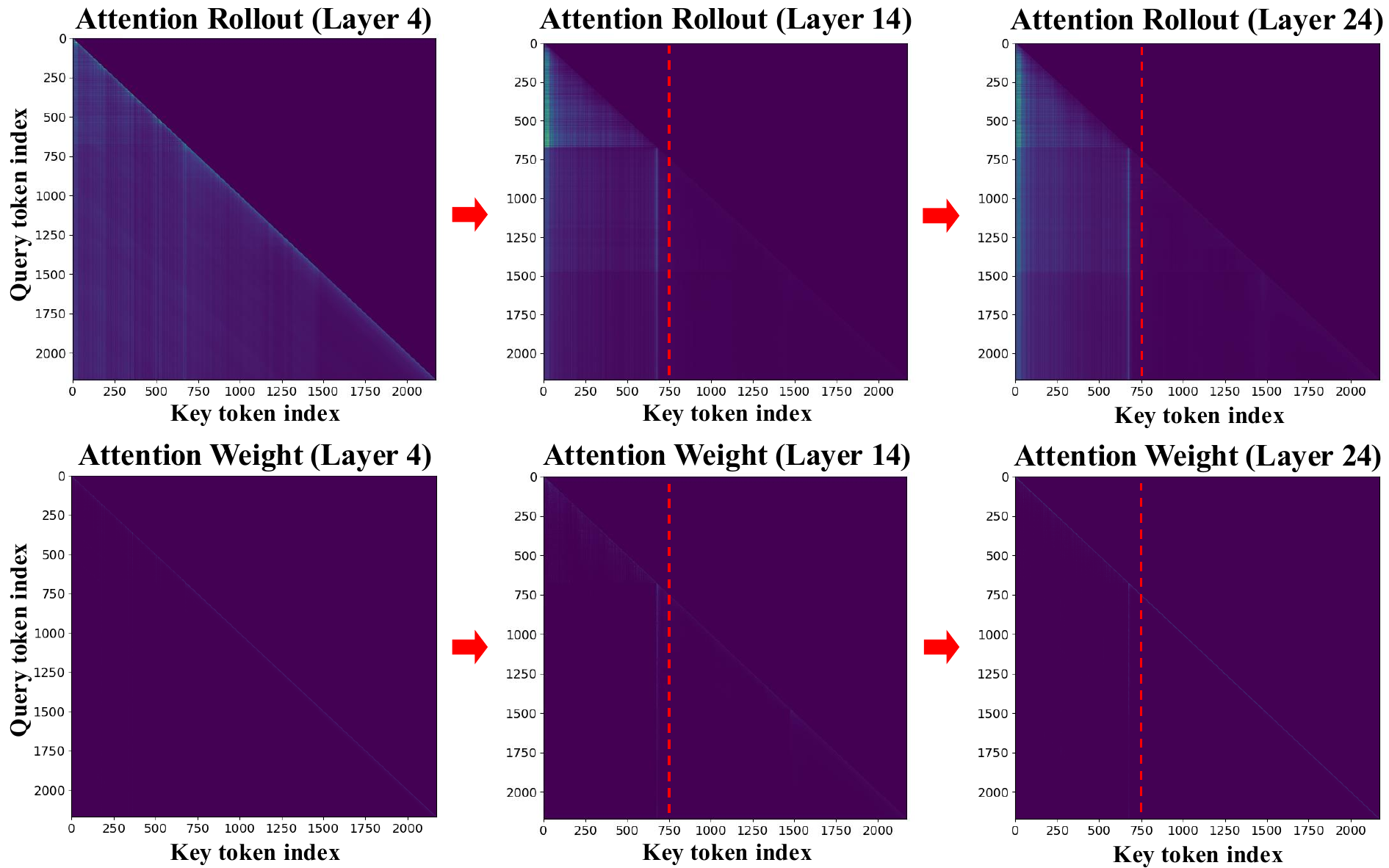}
  \vspace{-8mm}
  \caption{
\textbf{Attention rollout and weights across layers in VideoLLaMA2~\cite{cheng2024videollama}.} Attention rollout reveals a progressive focus on earlier tokens, stabilizing around the middle layers and persisting in deeper layers, whereas raw attention weights alone do not exhibit such a clear pattern.}
  \vspace{-5mm}
  \label{fig2}
\end{figure}
\subsection{FastAV}
\newpara{Analysis of audio-visual token importance via attention rollout.}
To explore the contribution of multimodal tokens across transformer layers, we use the attention rollout method~\cite{abnar-zuidema-2020-quantifying}, which tracks how information propagates through the network as depth of the layer increases. Attention rollout aggregates attention weights from the initial to the current layer. Since each transformer layer has multiple attention heads, we first aggregate them by averaging across heads, resulting in a single attention matrix $\mathbf{A}^l \in \mathbb{R}^{n \times n}$ for layer $l$, where $n$ is the number of tokens.
Each row of $\mathbf{A}^l$ represents a token’s attention distribution over all tokens. 
To preserve each token’s original representation and prevent the attention rollout from being dominated solely by raw attention weights, we incorporate residual connections by forming a convex combination of $\mathbf{A}^{l}$ and the identity matrix $\mathbf{I}$:
\begin{myequation}
\tilde{\mathbf{A}}^{\,l} = \alpha\,\mathbf{A}^{\,l} + (1-\alpha)\,\mathbf{I}.
\end{myequation}
Here, $\alpha$ plays the role of balancing residual connections and aggregated attention. Larger values highlight inter-token dependencies, while smaller values reinforce each token’s own representation, stabilizing the rollout. 
The cumulative attention rollout up to layer $l$ is computed as the sequential matrix multiplication of these modified attention matrices:
\begin{myequation}
\mathbf{R}^{\,l} = \tilde{\mathbf{A}}^{\,l}\,\tilde{\mathbf{A}}^{\,l-1}\cdots \tilde{\mathbf{A}}^{\,1}.
\end{myequation}
This matrix represents how much each token from the input layer influences every other token after propagating through all attention layers up to $l$.
\begin{figure*}[t]
  \centering
  \vspace{-2mm}
  \includegraphics[width=0.85\linewidth]{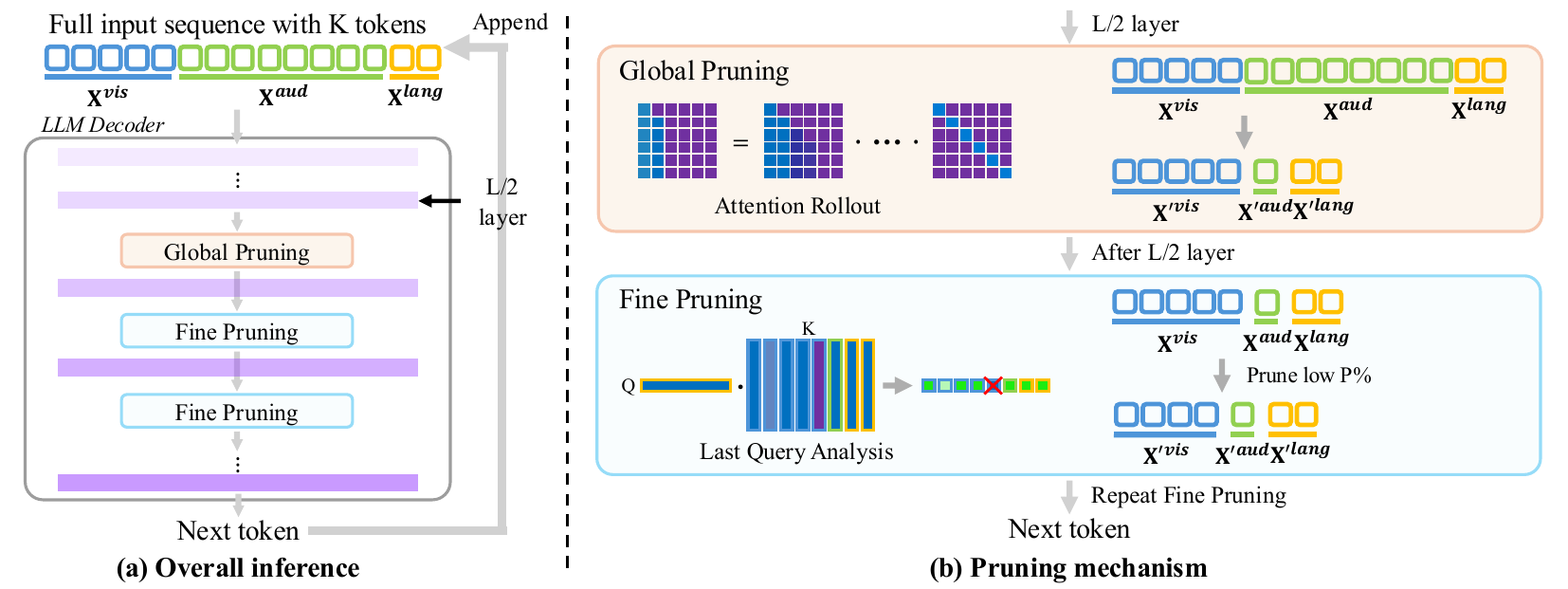}
  \vspace{-4mm}
  \caption{
   \textbf{Overview of the FastAV framework.} FastAV starts with the full input context and reduces computation through two-stage pruning. In the middle layer, global pruning removes later tokens guided by attention rollout. In subsequent layers, fine-grained pruning discards the least important P\% of remaining tokens based on last-query token analysis.
}
  \vspace{-5mm}
  \label{fig3}
\end{figure*}
We use attention rollout to gain deeper insight into the model’s behavior, rather than relying solely on raw attention weights. To illustrate its effectiveness, Fig.~\ref{fig2} compares attention rollout with raw attention weights at early (layer 4), middle (layer 14), and late (layer 24) stages of the 28-layer VideoLLaMA2~\cite{cheng2024videollama} network. In the first row, showing the attention rollout, we observe that attention gradually concentrates on the earliest tokens starting from the middle layer, as highlighted by the red line. By the late layer, this rollout pattern has stabilized. In contrast, the second row, showing raw attention weights, reveals no clear pattern. This highlights the effectiveness of attention rollout in capturing the model’s behavior and guiding global pruning decisions.

\newpara{Global pruning.} 
Guided by attention rollout, we first perform global pruning to remove tokens in later positions, as attention increasingly concentrates on earlier tokens after the middle layers (L/2), as shown in Fig.~\ref{fig3} (a). 
For both models, we analyze 100 non-test samples and apply an attention rollout threshold at the middle layer to remove less influential video and audio tokens, typically those occurring beyond position 750 in the sequence. As a result, approximately two-thirds of the later tokens are removed for VideoLLaMA2, while more than half are removed for video-SALMONN2.

\newpara{Fine pruning.} After global pruning, most tokens have already been discarded, so subsequent pruning must be applied carefully to further improve inference efficiency without harming performance.
We perform fine pruning in the layers following the middle layer where global pruning is applied, using the last query token in the attention weight, as illustrated in Fig.~\ref{fig3} (b). 
As in prior works~\cite{song2024hierarchical, jung2025avcd}, the last query token analysis efficiently identifies important tokens, as it directly influences the next token’s prediction without computing the full attention matrix. Let $\mathbf{Q}^l_{\text{last}} \in \mathbb{R}^{h \times 1 \times d}$ denote the last query feature at layer $l$, and $\mathbf{K}^l \in \mathbb{R}^{h \times n \times d}$ represents the key features of the $n$ tokens remaining after global pruning, where $h$ is the number of attention heads and $d$ is the hidden dimension.
The importance scores are computed as:
\begin{myequation}
\mathbf{s}^l = \text{mean}_{h} \Big(\text{softmax}(\mathbf{Q}^l_{\text{last}}  (\mathbf{K}^l)^\top) \Big),
\end{myequation}
where the mean is taken over attention heads. At each layer following global pruning, tokens with the lowest P\% scores are removed, iteratively refining the token set. 


\begin{table*}[t]
\centering
\renewcommand{\arraystretch}{0.9}
\caption{\textbf{Theoretical FLOPs and accuracy on VideoLLaMA2~\cite{cheng2024videollama} and video-SALMONN2~\cite{tang2025video} across AVQA~\cite{yang2022avqa}, MUSIC-AVQA~\cite{li2022learning}, and AVHBench~\cite{sung2024avhbench}.} FastAV significantly reduces FLOPs while maintaining comparable accuracy without additional training. NA indicates not applicable, as MUSIC-AVQA contains long videos unsuitable for video-SALMONN2.}
\label{tab:main_results}
\resizebox{0.83\linewidth}{!}{ 
\begin{tabular}{lccc|cc|cccc}
\toprule
\multirow{2}{*}{Method} & \multirow{2}{*}{\text{FLOPs$\downarrow$}} & \multirow{2}{*}{\text{Latency$\downarrow$}} & \multirow{2}{*}{\text{Memory$\downarrow$}}  
& \multirow{2}{*}{\text{MUSIC-AVQA$\uparrow$}}  & \multirow{2}{*}{\text{AVQA$\uparrow$}} 
& \multicolumn{3}{c}{AVHBench} \\
\cmidrule(lr){7-9}
& & & & & & \text{AV hallucination$\uparrow$} & \text{AV matching$\uparrow$} &\text{AV captioning$\uparrow$}\\
\midrule
{VideoLLaMA2~\cite{cheng2024videollama}}   & 100  & 0.43 & 22G     &  \textbf{81.3}   & 61.4       & 77.9           & 57.8             &  2.8            \\  w/ FastAV
                               & \textbf{56}  & \textbf{0.32} & \textbf{19G} &  {81.2}&  \textbf{62.3}     & \textbf{78.2}    & \textbf{69.0}             & \textbf{2.9}               \\ \midrule
{video-SALMONN2~\cite{tang2025video}} & 100  &0.44 & 28G   &  NA   & 57.6        &      64.5         &   \textbf{50.8}           &  \textbf{3.2}        \\ 
w/ FastAV
                               & \textbf{58} & \textbf{0.29} & \textbf{21G} &   NA       &\textbf{58.4}    &  \textbf{64.8}     &    50.7 &  3.1             \\
\bottomrule

\end{tabular}
}
\vspace{-5mm}
\end{table*}


\begin{table}[t]
\centering
\vspace{-3mm}
\renewcommand{\arraystretch}{0.9}
\caption{\textbf{Comparison of global pruning strategies with VideoLLaMA2 on AVHBench.} Low informative pruning performs best, highlighting the value of attention rollout.}
\label{tab:component_abl}
\resizebox{0.95\linewidth}{!}{%
\begin{tabular}{l|c|ccc}
\toprule
\multirow{2}{*}{Method} & \multirow{2}{*}{FLOPs} & \multicolumn{3}{c}{AVHBench} \\ \cmidrule(lr){3-5}
                        &                        & AV hallucination & AV matching & Avg \\ \midrule
 \rowcolor{lightgray} Vanilla                  &         100          & 77.9 & 57.8 & 70.7 \\
 Random                  &        \multirow{5}{*}{65}           & 77.2 & 54.2 & 69.0 \\
Top attentive           &  & 76.1 & 51.7 & 67.4 \\

Low attentive           &                     & 77.5 & 57.8 & 70.5 \\
Top informative           &  & 72.3 & 50.9 & 64.7 \\
Low informative (Ours)  &                     & \textbf{78.7} & \textbf{67.7} & \textbf{74.5} \\

\bottomrule
\end{tabular}
}
\vspace{-5mm}
\end{table}

\begin{table}[t]
\centering
\caption{\textbf{Comparison of fine pruning strategies with VideoLLaMA2 on AVHBench.} Low attentive pruning achieves the best performance, demonstrating its effectiveness.}
\label{tab4}
\resizebox{0.9\linewidth}{!}{%
\begin{tabular}{l|c|ccc}
\toprule
\multirow{2}{*}{Method} & \multirow{2}{*}{FLOPs} & \multicolumn{3}{c}{AVHBench} \\ \cmidrule(lr){3-5}
                        &                        & AV hallucination & AV matching & Avg \\ \midrule
 \rowcolor{lightgray} Vanilla                  &           100    & 77.9 & 57.8 & 70.7 \\
 Random                  &    \multirow{3}{*}{56}                      & 76.1 & 54.9 & 68.5 \\
Top attentive           &  & 74.5 & 52.8 & 66.8 \\
Low attentive (Ours)    &                     & \textbf{78.2} & \textbf{69.0} & \textbf{74.9} \\
\bottomrule
\end{tabular}
}
\vspace{-5mm}
\end{table}

\section{Experiments}
\label{sec:exp}

\vspace{-2mm}

\subsection{Experimental setup}
\vspace{-3mm}
\newpara{Baselines.}
We evaluate our approach using two representative AV-LLMs: VideoLLaMA2~\cite{cheng2024videollama} and video-SALMONN2~\cite{tang2025video}. 

\newpara{Datasets.}
AVQA~\cite{yang2022avqa} dataset contains 57,000 YouTube videos for real-world audio-visual understanding. MUSIC-AVQA~\cite{li2022learning} includes 45,867 question-answer pairs from 9,288 music videos, focusing on audio-visual reasoning. AVHBench~\cite{sung2024avhbench} evaluates audio-visual hallucinations and is divided into three subtasks: audio- or video-driven hallucination (AV hallucination), audio-visual matching (AV matching), and audio-visual captioning (AV captioning).

\newpara{Evaluation protocol.}
For AVHBench, we report accuracy excluding AV captioning. For AVQA, MUSIC-AVQA, and AV captioning, which have open-ended responses, we follow the GPT-assisted evaluation protocol from VideoLLaMA2\footnote{\tiny\url{https://github.com/DAMO-NLP-SG/VideoLLaMA2/tree/audio_visual}}
and report only the average score, as standard deviations are negligible. FLOPs are measured relative to the original theoretical one (set to 100) as in~\cite{chen2024image}, and latency indicates the time in seconds to generate a single token during a forward pass.

\newpara{Implementation details.}
For global pruning in VideoLLaMA2, all video tokens precede the audio tokens, so we keep only the first 10 audio tokens and prune the rest. In video-SALMONN2, video and audio tokens are interleaved at the frame level, so we prune the later frames while retaining the first 4. For fine pruning, we set the pruning ratio 
P to 20\% of the tokens at each layer after the middle layer.

\subsection{Experimental Results}
\newpara{Main results.} 
To demonstrate the effectiveness of FastAV, we conduct experiments on VideoLLaMA2 and video-SALMONN2 across three datasets: AVQA, MUSIC-AVQA, and AVHBench. As shown in Table~\ref{tab:main_results}, FastAV consistently reduces theoretical FLOPs by over 40\%, accelerates inference by approximately 30\%, and lowers memory consumption, all while maintaining strong performance across tasks. Notably, on VideoLLaMA2, accuracy on the AV matching task improves by over 10\%, suggesting that removing more than 99\% of audio tokens may enhance multimodal understanding.
FastAV significantly enhances computational efficiency, maintaining model performance without additional training.

\begin{figure}[t]
  \centering
 
  \includegraphics[width=0.95\linewidth]{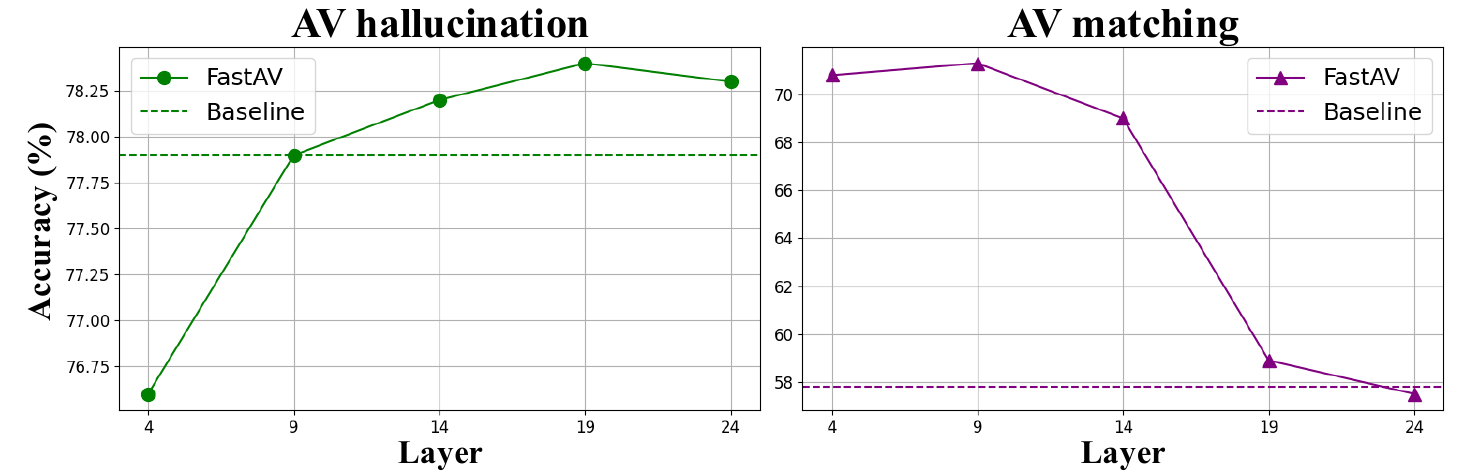}
  \vspace{-4mm}
  \caption{
\textbf{Layerwise accuracy of VideoLLaMA2 on AVHBench subtasks.} The middle layer are chosen to balance performance between AV hallucination and AV matching tasks.
}

   \vspace{-3mm}
  \label{fig4}
\end{figure}
\begin{table}[t]
\centering
\vspace{-2mm}
\renewcommand{\arraystretch}{0.9}
\caption{\textbf{Theoretical FLOPs and accuracy of VideoLLaMA2 under different pruning ratios P.} FastAV significantly reduces FLOPs as pruning increases, with a 20\% pruning ratio achieving the best accuracy at low FLOPs.}

\label{tab:per_abl}
\resizebox{0.75\linewidth}{!}{ 
\begin{tabular}{l|c|ccc}
\toprule
\multirow{2}{*}{P (\%)} & \multirow{2}{*}{FLOPs$\downarrow$} & \multicolumn{3}{c}{AVHBench} \\ \cmidrule(lr){3-5}
                                    &                        & AV hallucination & AV Matching & Avg \\ \midrule
0                                  & 65                & \textbf{78.7}  &     67.7  & 74.5     \\

10                                  & 59                   & 78.3   &     68.3  & 74.7     \\
20 (Ours)                                 & 56                    & 78.2   &     \textbf{69.0}   &\textbf{74.9}     \\
30                                  & \textbf{54}                    & 78.3      &    68.5    &74.8     \\
\bottomrule

\end{tabular}
}
\vspace{-3mm}
\end{table}

\newpara{Global pruning strategy.} To assess the effectiveness of our global pruning method, we compare several strategies on VideoLLaMA2 using AVHBench in Table~\ref{tab:component_abl} without applying fine pruning.
Vanilla represents the original model inference without pruning, and random pruning removes tokens arbitrarily while keeping FLOPs constant. In contrast, pruning top attentive tokens based on last-query attention weights significantly degrades performance, since high-attention tokens often carry critical information~\cite{song2024hierarchical, jung2025avcd}. Pruning low attentive tokens preserves vanilla performance, but our rollout-based low-informative strategy outperforms it, demonstrating that attention rollout captures token importance more effectively than raw attention for global pruning. Additionally, removing highly informative tokens leads to the worst performance, highlighting their essential role in our approach.

\newpara{Fine pruning strategy.}
Table~\ref{tab4} compares fine pruning strategies on VideoLLaMA2 using AVHBench. The low-attentive approach consistently outperforms random pruning, demonstrating its effectiveness as the second stage of FastAV.

\newpara{Pruning layer selection.}
To identify the optimal starting layer for two-stage pruning, we conduct layer-wise experiments on VideoLLaMA2 using AVHBench. Guided by attention rollout analysis, we select a middle layer (layer 14) as the pruning starting point.
As shown in Fig.~\ref{fig4}, pruning in the early layers degrades performance on the AV hallucination task, whereas pruning from the middle layer preserves performance and can even improve results across all tasks.



\newpara{Pruning percentage P.} To select an appropriate pruning ratio for fine pruning, we conduct experiments on VideoLLaMA2 using AVHBench. Table~\ref{tab:per_abl} reports results with pruning ratios increased in 10\% increments, showing that higher pruning ratios further reduce FLOPs. A 0\% pruning ratio corresponds to applying only global pruning, while a 20\% pruning ratio reduces FLOPs by approximately 44\% compared to the original inference and achieves the highest average performance.


\section{Conclusion}
We introduce FastAV, the first token pruning framework specifically designed for AV-LLMs, combining global pruning via attention rollout with fine pruning using last-query analysis to remove less informative tokens while preserving crucial information. Experiments show it reduces FLOPs by over 40\%, enabling efficient processing of long, complex multimodal inputs without degrading performance.

\vfill\pagebreak



\setstretch{0.96}
\setlength{\bibsep}{0.1ex}

\bibliographystyle{IEEEbib}
\bibliography{strings,refs}
\end{document}